\documentclass[5p, times]{elsarticle}
\makeatletter
\def\ps@pprintTitle{%
   \let\@oddhead\@empty
   \let\@evenhead\@empty
   \let\@oddfoot\@empty
   \let\@evenfoot\@oddfoot
}
\makeatother

\usepackage{bm}
\usepackage[svgnames,dvipsnames,table]{xcolor}
\usepackage[colorlinks]{hyperref}
\usepackage{amsmath}
\usepackage{amssymb}
\usepackage{adjustbox}
\usepackage{multirow}
\usepackage{tikz}
\usepackage{pgfplots}
\usepgfplotslibrary{patchplots}
\usetikzlibrary{trees,shapes,snakes,calc,matrix,positioning,fit,arrows,patterns,backgrounds,quotes,arrows.meta}
\usepackage{siunitx}
\usepackage{multicol}
\usepackage{caption}
\usepackage{subcaption} 
\usetikzlibrary{shapes.geometric}
\usetikzlibrary{shapes.arrows}
\usepgfplotslibrary{fillbetween}
\usepackage{array}
\usepackage{tabularx}
\usepackage{lipsum} 
 \usepackage{collcell}
\usepackage{hhline}
\usetikzlibrary{spy, calc}
\usepackage{array}
\usepackage{tabularx}
\usepackage{tabulary}
\usepackage{xspace}

\usepackage{bbding}
\usepackage{pifont}
\usepackage{diagbox}
\usepackage{booktabs}
\usepackage{makecell}
\usetikzlibrary{fit}
\usepackage{rotating}
\usepgfplotslibrary{colorbrewer}
\usepackage{fontawesome}
\usepackage{lscape}
\usepackage[para,online,flushleft]{threeparttable}
\usepackage{longtable}
\usepackage{tablefootnote}
\usepackage{makecell}
\usepackage{breakcites}
\usepackage{pifont}
\usepackage{hyphenat}
\begin{document}
\begin{frontmatter}

\title{Quick-CapsNet (QCN): A fast alternative to Capsule Networks}


\author[1]{Pouya Shiri}
\ead{pouyashiri@uvic.ca}
\author[1]{Ramin Sharifi}
\ead{raminsharifi@uvic.ca}
\author[1]{Amirali Baniasadi}
\ead{amiralib@uvic.ca}


\address[1]{Department of Electrical and Computer Engineering, University of Victoria, 3800 Finnerty Rd, Victoria, BC, V8P 5J2 Canada}

\begin{abstract}
The basic computational unit in Capsule Network (CapsNet) \cite{Sabour2017} is a capsule (vs. neurons in Convolutional Neural Networks (CNNs)). A capsule is a set of neurons, which form a vector. CapsNet is used for supervised classification of data and has achieved state-of-the-art accuracy on MNIST digit recognition dataset, outperforming conventional CNNs in detecting overlapping digits. Moreover, CapsNet shows higher robustness towards affine transformation when compared to CNNs for MNIST datasets. One of the drawbacks of CapsNet, however, is slow training and testing. 
This can be a bottleneck for applications that require a fast network, especially during inference. In this work, we introduce Quick-CapsNet (QCN) as a fast alternative to CapsNet, which can be a starting point to develop CapsNet for fast real-time applications. QCN builds on producing a fewer number of capsules, which results in a faster network. QCN achieves this at the cost of marginal loss in accuracy. Inference is 5x faster on MNIST, F-MNIST, SVHN and Cifar-10 datasets. We also further enhanced QCN by employing a more powerful decoder instead of the default decoder to further improve QCN.
\end{abstract}


\end{frontmatter}

\section{Introduction}
Deep Learning is a branch of Artificial Intelligence that imitates the human brain in data processing and pattern recognition. Deep Learning offers great performance across different computer vision tasks including image classification. Image classification can be challenging due to the ever-increasing complexity we witness in datasets \cite{Sabour2017}. One set of the most prevalent deep learning approaches to classifications are those based on Convolutional Neural Networks (CNNs). The word “convolutional” reflects the Convolutional Layer, which is the fundamental layer in CNNs.  This layer convolves the weights (called filters) with the input to create the output.
\par
CNNs consist of several convolutional layers followed by several fully-connected layers \cite{Zhao2017a}. The advancement in CNN research is due to the availability of extensive training data and advanced hardware. CNNs have achieved exceptional performance in classification. The main breakthrough of CNN was AlexNet \cite{Krizhevsky} and later GoogLeNet \cite{Szegedy2015}, winners of the ImageNet classification competition (ILSVRC) \footnote{ImageNet competitions include several challenges and one of them is image classification. In this challenge, there are many images of many different categories (normally 1000 categories) to be classified. The success of AlexNet, GoogLeNet and VGG in ImageNet competitions is a strong indication that CNNs are great choices for classification purposes.}
\par
CNN classifiers are networks consisting of several layers, each of which produces an output called a feature map. Starting from the very first layer, CNNs aim to extract features from the input images and make the extracted information more meaningful as the network proceeds to the next layers.
\par
One particular type of layer in CNNs is the Pooling Layer. The pooling layer reduces the size of the feature map so that there are fewer computations and trainable parameters in the network. Practically, this reduction means taking a smaller number of values in the input feature map and then eventually replacing them with a single number (e.g., maximum) to create a smaller feature map. The size reduction comes with information loss \footnote{CapsNet avoids using the pooling layer to protect against this loss of information.}.
\par
One of the drawbacks associated with using pooling layers in CNNs is that they do not consider the relationship between low-level and high-level features. A high-level feature like a face consists of low-level features such as nose, eyes, lips and so on. The spatial relationship of these low-level features is not considered by CNN. Consequently, a face with its eyes replaced with its lips can still be classified as a face.  
\par
 CapsNet was designed to address the above drawback \cite{Sabour2017}. Since its introduction, CapsNet has been a hot research topic \cite{Xiang2018,Mobiny2018,Kim2018a}. A CapsNet is a network that uses capsules as the basic unit of computation (in CNNs, the basic unit is a neuron). Although the concept was introduced earlier \cite{Hinton2011}, there was no practical training mechanism for a network using capsules. In \cite{Sabour2017} the authors proposed an algorithm called “Routing by Agreement”, or “Dynamic Routing” (DR) to make training capsules possible. Unlike CNNs, CapsNet considers the relationship between low-level and high-level features using the DR algorithm. 
 \par
In this work, we investigate if there is room for improvement in CapsNet’s speed. To this end, we propose a variant of CapsNet, which is faster but competitively accurate. In conventional CapsNet, early layers use convolutional layers for extracting features and transforming image samples to a set of activation tensors, which in turn are converted to primary capsules (PCs). We propose Quick-CapsNet (QCN), in which we redesign this conventional feature extraction method.  QCN generates much fewer PCs compared to CapsNet and achieves faster training and testing. Moreover, QCN maintains robustness to affine transformations applied to the input images. We also change the architecture of the decoder in the network and use a more powerful decoder. This alternative decoder consists of deconvolution layers and uses a different method for masking the output vector. 
\par
The rest of the paper is organized as follows. Section II presents related works. Section III focuses on the motivation of our work. Background and methodology are explained in section IV. Section V reports experimental results. We offer concluding remarks in section VI.

\section{Related Works}
Since CapsNet was introduced, there have been several studies that aim at making it more computationally efficient and effective. Some studies have made CapsNet faster in terms of the time of convergence or training and inference time while others focused on the testing accuracy of the network. M. do Rosario et al. proposed Multi-Lane Capsule Networks (MLCN) \cite{DoRosario2019} as an energy-efficient variant of CapsNet. MLCN  provides two-fold faster training and inference.  It contains several data-independent lanes, each of which creates a dimension of the digit capsules. The lanes are similar to the baseline CapsNet in terms of architecture. The independent construction of capsules allows for parallel execution, improving performance. This work, however, does not investigate robustness to affine transform. 
\par
Rajasegaran et al. \cite{Rajasegaran2019} proposed DeepCaps includeing  a deep version of CapsNet. DeepCaps makes two major contributions:
\begin{enumerate}
    \item A new dynamic routing algorithm based on 3D convolution
    \item A class-independent decoder
\end{enumerate}
DeepCaps outperforms CapsNet for Cifar-10, Fashion-MNIST and SVHN datasets. In addition, it reduces the number of parameters significantly. In this work, we make some modifications to the class-independent decoder introduced in \cite{Rajasegaran2019} and employ it in QCN. 
\par
Zuo et al. \cite{Zou2018} proposed a new activation function as a replacement for the current “squash” activation employed by CapsNet. Their proposal improves the convergence speed of the network. Moreover, they added “least weight loss” to the loss function to increase the generalization of the network and network accuracy. 
Ahmed et al. introduced Star-Caps \cite{Ahmed} to address the high computational complexity of CapsNet. Their solution works better for small scale images. They used a new algorithm instead of dynamic routing. Their method makes a binary decision whether to route between two capsules or not. This leads to better performance in terms of speed and stability. In Self-routing Capsule Networks \cite{Hahn}, Hahn et al. replaced dynamic routing with a faster  alternative. In contrast to dynamic routing, self-routing does not need an agreement between capsules. The authors presnt their solution as the Mixture-of-Experts method. Bahadori \cite{TahaBahadori} proposed an eigendecomposition method for faster convergence of prediction vectors.

\section{Motivation}
Even though CapsNet has advantages over conventional CNNs, it is slow. This is due to the newly added structure of neurons as vectors and the dynamic routing algorithm. In this work, we focus on making CapsNet faster, while maintaining its advantages. In this section, we explain why we chose to focus on reducing the number of capsules to make CapsNet faster. 
\par
The architecture of CapsNet is explained in the next section in detail. CapsNet includes a feature extractor, which creates the first capsules that are further processed in the subsequent capsule layers. These capsules are called primary capsules (PCs). The number of PCs is one of the parameters that directly impacts the network complexity and speed. The higher the number of PCs, the more weights are required. This could be explained by the weights assigned to the affine transform matrix multiplication (details explained in the next section). Moreover, the dynamic routing algorithm \cite{Sabour2017}, becomes more expensive as the number of PCs increase. However, network accuracy is expected to decline as the number of PCs decrease. This can be explained by the fact that each additional PC contributes to creating a more representative model and improves generalization ability. In the meantime, this adds to network complexity.
\par
Our experiments verify the effect of the number of PCs on network performance. We evaluated the performance of the network in terms of accuracy, number of parameters and the speed i.e. training and testing time for different PC numbers for the Cifar-10 dataset. 
\par

\begin{figure}[htp]
    \centering
    \includegraphics[width=\linewidth]{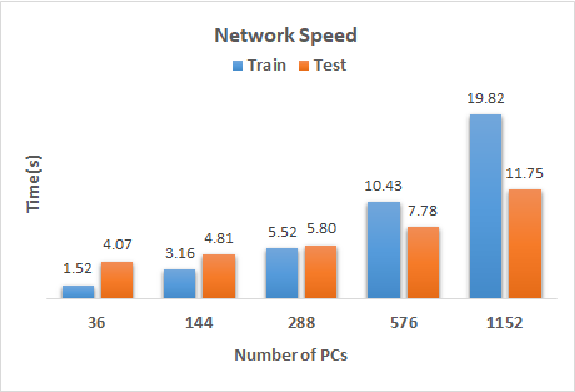}
    \caption{Exploring the effect of the number of PCs on the network speed for Cifar-10 dataset. Training time is divided by 10. Note that network becomes slower as the number of PCs increase.}
    \label{fig:motiv_1}
\end{figure}

Figure \ref{fig:motiv_1} shows the effect of the number of PCs on the network speed. As expected, the network becomes slower when increasing the number of PCs. Increasing the number of PCs from 36 to 1152, results in 13x slower training and 2.9x slower testing time.

\begin{table}[htp]
\caption{The network accuracy and the number of parameters for different numbers of PCs. The network accuracy drops as the number of PCs decrease. The number of PCs has a significant effect on the number of parameters.} 
\centering 
\resizebox{\linewidth}{!}{\begin{tabular}[width=\linewidth]{c|c c c c c} 
\hline\hline 
\textbf{\#PC} & PC=36 & PC=144 & PC=288 & PC=576 & PC=1152 \\ [0.5ex] 
\hline 
\textbf{Accuracy(\%)} & 48.45 & 62.24 & 65.68 & 67.96 & 69.34 \\ 
\hline
\textbf{\#Parameters} & 4,066,824 & 4,810,272 & 5,801,536 & 7,784,064 & 11,749,120 \\ [1ex] 
\hline 
\end{tabular}}
\label{table:tmotiv_1} 
\end{table}

Table \ref{table:tmotiv_1} shows the effect of the number of PCs on the number of parameters and network accuracy. As the table shows, network accuracy drops as the number of PCs decrease. The number of PCs has a significant effect on the number of parameters.
\par
The experiments showed that having fewer capsules, results in a faster network. However, the current feature extractor of CapsNet loses accuracy by lowering the number of PCs. In this work, we propose a different feature extraction method to produce fewer capsules while maintaining accuracy.

\section{Background and Methodology}

\begin{figure*}
    \centering
    \includegraphics[width=\textwidth,height=\textheight,keepaspectratio]{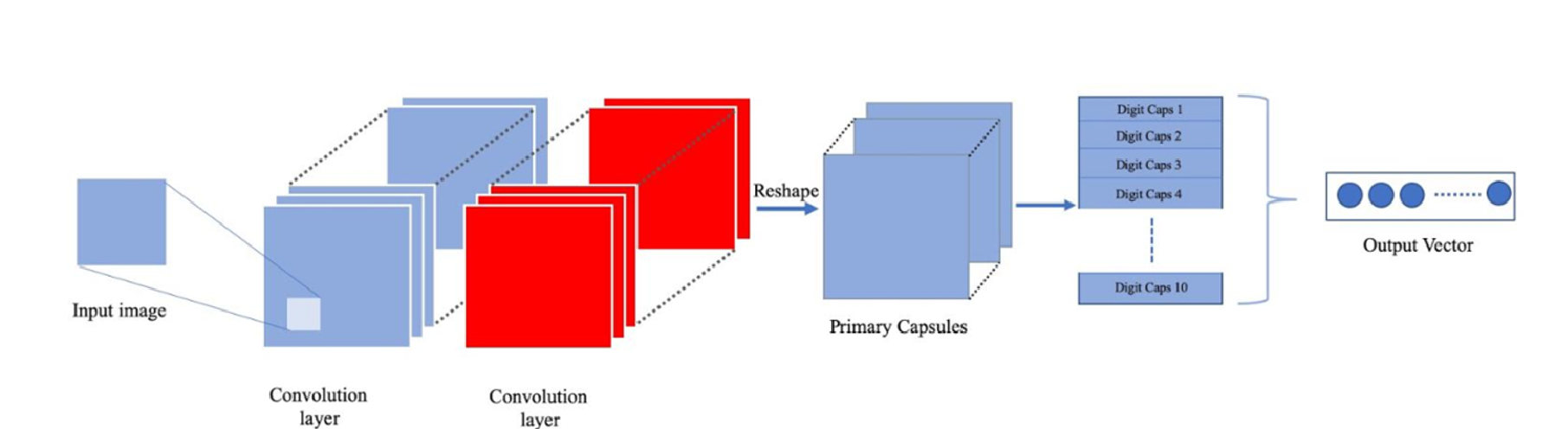}
    \caption{CapsNet architecture. The input image goes through two convolution layers. Then and after being reshaped, it enters the primary capsule layer. At the end, the output vector will be the largest magnitude of the vectors present in digit caps.}
    \label{fig:method_1}
\end{figure*}

Figure \ref{fig:method_1} shows the baseline architecture proposed for CapsNet \cite{Sabour2017}. As the figure shows, the network begins with two convolutional layers that extract the low-level features of the input image. The output of the second convolutional layer is reshaped into 8D vectors. These 8D vectors are referred to as Primary Capsules (PCs). There could be several more capsule layers once the first layer of capsules are formed.

\begin{figure}[htp]
    \centering
    \includegraphics[width=\linewidth]{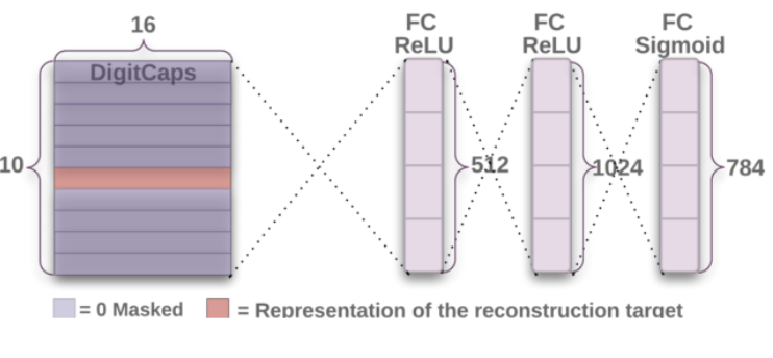}
    \caption{CapsNet Reconstruction Sub-network \cite{Sabour2017}. Images of CapsNet are reproduced to create a new term in the loss.}
    \label{fig:method_2}
\end{figure}

CapsNet includes a reconstruction network, which works as a decoder to reconstruct the input images. As figure \ref{fig:method_2} shows, to make the reconstruction network, the output of the final layer of capsules is connected to 3 FC layers. The reconstructed images are then compared with the input images to add a Reconstruction Loss term. The Loss term is added to margin loss to build the total loss function of CapsNet.

In the original CapsNet implementation (which we call baseline CapsNet) \cite{Sabour2017} there is just one capsule layer. The output this capsule layer is multiplied with a matrix (referred to as the affine transform stage). This is where the robustness to affine transformation takes place.  The result of this multiplication is known as the capsule predictions of the current layer. The final capsule layer contains a number of 16D vectors. The number of these vectors corresponds to the number of categories in the classification task. The length of these vectors determines the probability of the input image belonging to a specific class and the angle of vectors represent various instantiation parameters such as width and scale of an entity. A long vector on a top-level capsule, means the input image is more likely related to the category corresponding to that capsule.  
\par
The relationship between capsule layers is handled by the dynamic routing (DR) algorithm. All capsules in the previous layer are connected in a fully-connected manner to all capsules in the next layer. DR algorithm determines the weights of the connection dynamically. It takes capsules from the previous layer and comes up with an agreement (agreeing on predictions for each class) among the predictions of these capsules and the capsules of the next layer iteratively in each forward pass.
\par
The loss function used for CapsNet consists of two parts: margin loss and reconstruction loss. Margin loss relies on increasing the loss value when predictions are not correct. There is a separate term for each capsule in margin loss.  The term for each capsule is as follows:
\[ L_k = T_k \max(0,m^+-||v_k||)^2 + \lambda(1-T_k)\max(0,||v_k||-m^-)^2 \]

Where $T_k$ is 1 when the entity of class k is present and 0 otherwise, $||v_k||$ is the length of k-th capsule, $\lambda$ is the down-weighting factor for classes that are not present (0.5 is a reasonable choice),  and $m^-$ and $m^+$ are used so that classes with very high or very low probability do not affect the loss function. Their values are taken 0.9 and 0.1 respectively.

\par

The number of PCs directly determines the computation complexity. The affine transform stage and dynamic routing become more computationally expensive as the number of PCs increases. Reducing the number of PCs in the baseline architecture, however, can reduce accuracy. Our goal is to reduce the number of PCs while maintaining accuracy.
\par
\begin{figure*}
    \centering
    \includegraphics[width=\textwidth,height=\textheight,keepaspectratio]{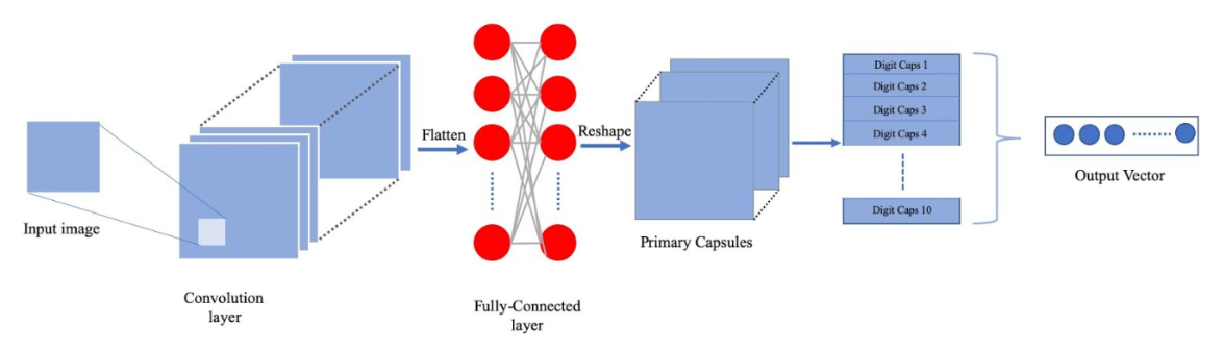}
    \caption{QCN architecture. Compared to the baseline, notice that the second Convolution layer is replaced with an FC layer.}
    \label{fig:method_3}
\end{figure*}

In this work we replace the second convolution layer with a Fully-Connected (FC) layer. Intuitively, this translates to including the contribution of all neurons in the output feature map of the convolutional layer. FC layer builds a representation that summarizes all of the neurons in the previous layer. Figure \ref{fig:method_3} shows the modified architecture. As presented, we feed the output of the first convolutional layer to an FC layer. The output of the FC layer is reshaped to create PCs, which are the inputs for the dynamic routing algorithm. 

In the baseline CapsNet, there are 1152 8D PCs. We chose to minimize the number of outputs of the FC layer for two reasons. The first reason has to do with the intrinsic feature of FC layers. Since the number of parameters in FC layers depends on the number of inputs and outputs. Moreover, the input of the FC layer already has a significantly large number of neurons (being the output of a convolutional layer). Therefore we aim at having as few outputs as possible. This is to avoid creating a heavy network in terms of number of parameters. Secondly, as we presented in the motivation section, the fewer PCs we have, the faster the network becomes. However, there is a minimum number of PCs needed to achieve an acceptable level of accuracy. Based on our experiments,  we found this number to be 4 for QCN. 

 We change the default decoder of CapsNet to a more powerful one, a class-independent. We refer to the network with the alternative decoder as QCN+. This decoder disregards the FC layers used in the default decoder and integrates deconvolution layers instead. To this end, the output vectors of CapsNet are fed to consecutive deconvolution layers of differenet kernel sizes. A deconvolution layer is better at capturing spatial relationships compared to FC layers for reconstruction. The other advantage of using deconvolution instead of FC layers is that it includes fewer parameters, as it comes with the weight sharing property. 
\par
Apart from using deconvolution, the class-independent decoder has a different input compared to the original decoder. The original decoder, builds output vectors of CapsNet and feeds them to the reconstruction network. During training, the class-independent decoder builds the output vectors based on the ground truth labels. In other words, all the output vectors are masked (zeroed out) except for the vector corresponding to the ground truth label. During testing, all the vectors are masked except the vector with the largest activity (length). This method is class-dependent,as there is a different distribution for each dimension. The alternative approach used here is to disregard the values instead of masking them. In other words, only one vector is kept and fed to the decoder during training and testing. This method is class-independent as for each dimension of the output vector, there is a single joint distribution among all categories .

\section{Experiments and Results}
In this section, we explain the experiments and report the results. We tested QCN on the following small-scale datasets: MNIST \cite{LECUN}, F-MNIST \cite{Xiao2017}, Cifar-10 \cite{Krizhevsky2009}, SVHN \cite{Netzer2011} and affine-transformed MNIST (Aff-NIST) datasets. CapsNet cannot yet be tested against large-scale datasets such as ImageNet , as increasing the number of categories increases the training time and the network size significantly. Note that since CapsNet is still at its early stages, innovative solutions are verified using the above datasets.
\par
\begin{table}[htp]
\caption{Datasets used in this work and their properties.} 
\centering 
\resizebox{\linewidth}{!}{\begin{tabular}[width=\linewidth]{|c|c| c| c| c|c|} 
\hline \hline
\textbf{Name} & \textbf{Image Size} & \textbf{\#Channels} & \textbf{Training samples} & \textbf{Test Samples} & \textbf{Baseline Acc.(\%)} \\ [1ex]

\hline
MNIST & 28x28 & 1 & 50,000 & 10,000 & 99.47\% \\ [1ex]
\hline
F-MNIST & 28x28 & 1 & 50,000 & 10,000 & 89.97\% \\ [1ex]
\hline
SVHN & 32x32 & 3 & 73,257 & 26,032 & 91.42\% \\ [1ex]
\hline
CIFAR-10 & 32x32 & 3 & 50,000 & 10,000 & 68.33\% \\ [1ex]

\hline 
\end{tabular}}
\label{table:table_dsets} 
\end{table}

Table \ref{table:table_dsets} summarizes the different datasets we used in our experiments and their features. The baseline accuracy column reports  the test accuracy of basline CapsNet implementation. MNIST and Fashion-MNIST (F-MNIST) datasets share the same data format. They both contain 28x28 grey-scale images and have 50,000 and 10,000 samples in training and testing sets, respectively. In addition they contain samples of 10 different classes. The difference between the two is that MNIST contains images of handwritten digits while F-MNIST includes samples of different pieces of clothing and is therefore more complex. 
\par
Cifar-10 and SVHN datasets both contain 32x32 RGB images of 10 different classes. Cifar-10 and SVHN include 50,000 and 10,000 (Cifar-10) and 73,257 and 26,032 (SVHN) samples in training and testing sets. SVHN includes images of the numbers of different houses. Each digit in these images is cropped to create a single-digit image in training and testing sets. Cifar-10 is the most complex dataset among the four. It contains 10 almost non-related categories i.e. airplanes, cars, birds, cats, deer, dogs, frogs, horses, ships, and trucks. The variation in the background is another factor that makes this dataset more challenging than the other three datasets.

\par

Here we compare QCN, QCN+ and CapsNet in terms of accuracy, the number of parameters and network speed. We use MNIST, F-MNIST, SVHN, Cifar-10 and AffNIST datasets. We implemented QCN using the PyTorch implementation of CapsNet\footnote{https://github.com/gram-ai/capsule-networks}. In all our experiments we use a 2080Ti GPU. Training is performed for 50 epochs and experiments are repeated five times. As there was little variance among the results of the experiments, we report the average. 
\par
We run our experiments on four datasets including MNIST, F-MNIST, SVHN and CIFAR-10. We evaluate two architectures: QCN and QCN+. We report for different number of generated PCs (4, 6 or 8 PCs). We also report the testing accuracy and the number of parameters for all experiments. Then we compare the speed of QCN and QCN+ with the baseline CapsNet for six PCs.
\par
 As mentioned in previous sections, there are 1152 PCs in the original implementation. We reduce this number to 4, 6 and 8 PCs. The upper limit is set to 8 to avoid adding too many parameters to the baseline implementation.
The lower limit is set to 4 because for fewer numbers of PCs. We have observed that further reductions reduces accuracy significantly.

\subsection{Network Speed-Up}
\begin{figure}
    \centering
    \includegraphics[width=\linewidth]{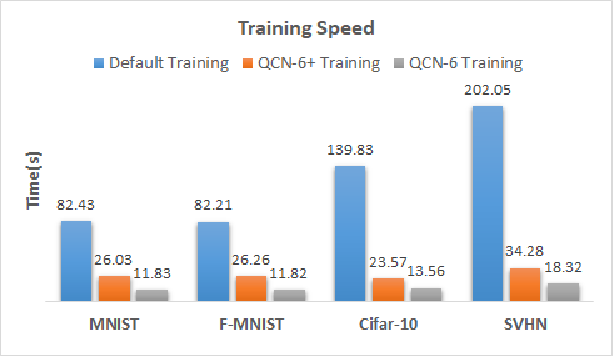}
    \caption{Network training speed in QCN and QCN+ compared to the baseline. The training time is shown for 4 datasets. QCN is significantly faster in training compared to the baseline CapsNet. QCN+ is slower than QCN due to the use of deconvolution layers.}
    \label{fig:result_1}
\end{figure}

We measured training and testing times for three different number of PCs mentioned above (4,6 and 8). As there was a very little variation in network speed for different number of PCs, we only report for QCN-6 (QCN with 6 PCs) and QCN-6+ (QCN+ with 6 PCs). Figure \ref{fig:result_1} shows the network training times for MNIST, FMNIST, SVHN and Cifar-10 datasets. As the figure shows, in training, QCN and QCN+ are 10x and 6x faster than the baseline for Cifar-10 and SVHN datasets, respectively. During inference, QCN and QCN+ are 7x and 3x faster on MNIST and F-MNIST datsets, respectively.

QCN+ is slower than QCN in most cases, as using multiple deconvolution layers results in more computations compared to employing FC layers.

\begin{figure}
    \centering
    \includegraphics[width=\linewidth]{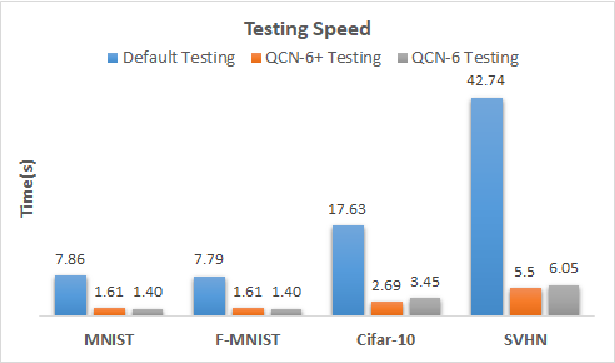}
    \caption{Network testing speed in QCN and QCN+ compared to the baseline. The training time is shown for four datasets under default CapsNet and QCN and QCN+. QCN is significantly faster than the baseline CapsNet in testing. }
    \label{fig:result_2}
\end{figure}

Figure \ref{fig:result_2} shows the network testing times for MNIST, FMNIST, SVHN and Cifar-10 datasets. As the figure shows, for both QCN and QCN+, testing is nearly 5x and 7x faster on SVHN and Cifar-10 datasets, respectively. QCN and QCN+ are 5.5x faster for MNIST and FMNIST datasets.


\subsection{Number of Parameters and Accuracy}
The number of parameters in QCN depends on the number of PCs. Table \ref{table:tres_param} and \ref{table:tres_acc} show the number of parameters and the change in the network accuracy for all four datasets. 
As the table \ref{table:tres_param} shows, QCN and QCN+ are lighter networks compared to the baseline, as they come with a smaller number of parameters. Table \ref{table:tres_acc} shows that there is loss of accuracy in all cases. This loss is marginal for some cases. With 8 PCs, QCN and QCN+ achieve the highest accuracy. In this case, QCN has 2.6\% and 7.2\% fewer parameters compared to the baseline for MNIST/FMNIST and Cifar-10/SVHN, respectively. QCN+ has even fewer number of parameters as it shows 16.5\% reduction for all datasets. QCN+ provides a marginally better accuracy compared to QCN for almost all cases. This is the result of employing a powerful decoder. With 8 PCs, QCN and QCN+ lose 0.25\% and 0.19\% accuracy for MNIST, 1.17\% and 1.13\% for F-MNIST. The change of decoder does not impact MNIST and F-MNIST significantly as input images are very simple and employ only one channel. Consequently even a conventional decoder performs well on reconstructing images. This is different for more complex datasets such as Cifar-10 and SVHN. QCN and QCN+ show 4.16\% and 1.16\% accuracy loss on Cifar-10, and 6.45\% and 5.22\% loss on SVHN. 


\begin{table}[htp]
\caption{Comparing the accuracy between QCN, QCN+ and the baseline CapsNet. QCN+ achieves higher accuracy.} 
\centering 
\resizebox{\linewidth}{!}{\begin{tabular}[width=\linewidth]{|c|c| c| c| c|} 
\hline\hline 
\textbf{Dataset} & MNIST & F-MNIST & Cifar-10 & SVHN \\ [0.5ex] 
\hline 
\textbf{Baseline Acc.(\%)} & 99.47 & 89.97 & 68.33 & 91.42 \\ 
\hline
\textbf{QCN Acc. (4 PCs)} & 99.17 & 88.63 & 63.12 & 84.55 \\ [1ex]
\hline
\textbf{QCN+ Acc. (4 PCs)} & 99.22 & 88.12 & 65.55 & 85.34 \\ [1ex]
\hline
\textbf{QCN Acc. (6 PCs)} & 99.29 & 88.99 & 64.28 & 85.51 \\ [1ex]
\hline
\textbf{QCN+ Acc. (6 PCs)} & 99.19 & 88.76 & 66.12 & 85.01 \\ [1ex]
\hline
\textbf{QCN Acc. (8 PCs)} & 99.22 & 88.80 & 64.18 & 84.97 \\ [1ex]
\hline
\textbf{QCN+ Acc. (8 PCs)} & 99.28 & 88.84 & 67.18 & 86.20 \\ [1ex]

\hline 
\end{tabular}}
\label{table:tres_acc} 
\end{table}

\begin{table}[htp]
\caption{Comparing the number of parameters between QCN, QCN+ and the baseline CapsNet. QCN+ includes fewer number of parameters.} 
\centering 
\resizebox{\linewidth}{!}{\begin{tabular}[width=\linewidth]{|c|c| c| c| c|} 
\hline\hline 
\textbf{Dataset} & MNIST & F-MNIST & Cifar-10 & SVHN \\ [0.5ex] 
\hline
\textbf{Baseline \#Params} & \multicolumn{2}{|c|}{8.21M} & \multicolumn{2}{|c|}{11.75M} \\ [1ex]
\hline
\textbf{QCN \#Params (4 PCs)} & \multicolumn{2}{|c|}{4.71M} & \multicolumn{2}{|c|}{8.54M} \\ [1ex]
\hline
\textbf{QCN+ \#Params (4 PCs)} & \multicolumn{2}{|c|}{3.59M} & \multicolumn{2}{|c|}{5.09M} \\ [1ex]
\hline
\textbf{QCN \#Params (6 PCs)} & \multicolumn{2}{|c|}{6.35M} & \multicolumn{2}{|c|}{13.26M} \\ [1ex]

\hline
\textbf{QCN+ \#Params (6 PCs)} & \multicolumn{2}{|c|}{5.23M} & \multicolumn{2}{|c|}{7.45M} \\ [1ex]
\hline
\textbf{QCN \#Params (8 PCs)} & \multicolumn{2}{|c|}{7.99M} & \multicolumn{2}{|c|}{10.90M} \\ [1ex]
\hline
\textbf{QCN+ \#Params (8 PCs)} & \multicolumn{2}{|c|}{6.87M} & \multicolumn{2}{|c|}{9.81M} \\ [1ex]

\hline 
\end{tabular}}
\label{table:tres_param} 
\end{table}

\subsection{Robustness to Affine Transformations}
We investigated how QCN impacts affine robustness. Measuring robustness to affine transformations is critical, as it is one of the main advantages of CapsNet over conventional CNNs. As such, any modification to CapsNet has to make sure that affine robustness is protected. To this end, we trained CapsNet and QCN on 28x28 images of MNIST dataset centered on a 40x40 grid and tested the network on the affine transformed MNIST. Baseline and QCN show 39.7\% and  29.9\% accuracy. QCN+ also achieves the same accuracy as the result of the simple one-channel input images in MNIST. This accuracy loss for QCN comes with  8.91x and 6.08x faster training and testing.

\section{Conclusion and Discussion}
CapsNet is still at its early stages.  This work improves CapsNet's etwork speed by proposing a modified architecture referred to as Quick-CapsNet (QCN). QCN spends significantly less time in training and testing. QCN comes with a slight drop in the testing accuracy. We also introduce QCN+, an enhanced extension of QCN, equipped with a more powerful decoder. QCN+ has fewer number of parameters compared to QCN and provides higher. Applications requiring real-time fast inference benefit from QCN and QCN+. 
\par

\section*{Acknowledgment}
This research has been funded in part or completely by the Computing Hardware for Emerging Intelligent Sensory Applications (COHESA) project. COHESA is financed under the National Sciences and Engineering Research Council of Canada (NSERC) Strategic Networks grant number
NETGP485577-15.



\bibliographystyle{unsrt}


\end{document}